\documentclass[lettersize,journal]{IEEEtran}
\usepackage{amsmath,amsfonts}
\usepackage{algorithmic}
\usepackage{algorithm}
\usepackage{array}
\usepackage[caption=false,font=normalsize,labelfont=sf,textfont=sf]{subfig}
\usepackage{textcomp}
\usepackage{stfloats}
\usepackage{url}
\usepackage{verbatim}
\usepackage{graphicx}
\usepackage{cite}
\usepackage[hidelinks]{hyperref}
\usepackage{bm}
\usepackage{siunitx}
\usepackage[inline]{enumitem}

\begin{document}

\title{Simplifying ROS2 controllers with a modular architecture for robot-agnostic reference generation}

\author{%
Davide Risi, 
Vincenzo Petrone,~\IEEEmembership{Member,~IEEE,}
Antonio Langella,
Lorenzo Pagliara,~\IEEEmembership{Student Member,~IEEE,}
Enrico Ferrentino,~\IEEEmembership{Member,~IEEE,}
Pasquale Chiacchio
\thanks{
Authors are with the Department of Information Engineering, Electrical Engineering and Applied Mathematics (DIEM), University of Salerno, 84084 Fisciano, Italy (email: \{drisi, vipetrone, alangella, lpagliara, eferrentino, pchiacchio\}@unisa.it).
}
\thanks{
\textcopyright 2026 IEEE.
Personal use of this material is permitted.
Permission from IEEE must be obtained for all other uses, in any current or future media, including reprinting/republishing this material for advertising or promotional purposes, creating new collective works, for resale or redistribution to servers or lists, or reuse of any copyrighted component of this work in other works.
}
}

\markboth{Accepted version, published at https://doi.org/10.1109/RAP.2026.3732617 --- IEEE Robotics and Automation Practice}{}

\IEEEpubid{0000--0000/00\$00.00~\copyright~2021 IEEE}

\maketitle

\begin{abstract}
This paper introduces a novel modular architecture for ROS2 that decouples the logic required to acquire, validate, and interpolate references from the control laws that track them. 
The design includes a dedicated component, named \emph{Reference Generator}, that receives references, in the form of either single points or trajectories, from external nodes (e.g., planners), and writes single-point references at the controller's sampling period via the existing \texttt{ros2\_control} chaining mechanism to downstream controllers.
This separation removes duplicated reference-handling code from controllers and improves reusability across robot platforms. 
We implement two reference generators: one for handling joint-space references and one for Cartesian references, along with a set of new controllers (PD with gravity compensation, Cartesian pose, and admittance controllers) and validate the approach on simulated and real Universal Robots and Franka Emika manipulators. 
Results show that (i) references are tracked reliably in all tested scenarios, (ii) reference generators reduce duplicated reference-handling code across chained controllers to favor the construction and reuse of complex controller pipelines, and (iii) controller implementations remain focused only on control laws. 
\end{abstract}


\section{Introduction}
\IEEEPARstart{T}{he} ROS2 framework aims to support an ecosystem where researchers and industries can rapidly set up robotic systems by reusing and composing open-source community-developed components \cite{macenski_robot_2022}.
Sharing this vision, the \texttt{ros2\_control} framework adopts design principles such as hardware-interface abstraction and modular design promoting controller reusability across robots and applications~\cite{Chitta2017,ros2-control}.
Notably, the controller chaining mechanism has been introduced to enable users to build complex behaviors from reusable components~\cite{ros2-controller-chaining}.
Its modular design enables inter-operability among control modules developed by different actors.
It also simplifies controller development by moving resource and lifecycle management out of individual controllers.

While this design successfully decouples controllers from hardware, it is less successful in decoupling control logic from input processing.
In contrast to the Single Responsibility Principle (SRP)~\cite{Martin2008}, standard controllers like the \texttt{AdmittanceController} and \texttt{JointTrajectoryController} \cite{ros2-controllers} act as monolithic entities that handle both reference processing and loop closure.
This leads to code duplication and undermines reusability.
In turn, it increases development time and maintenance costs, with a steeper learning curve for new developers.
Potentially, this could hinder the adoption of \texttt{ros2\_control} in educational and research contexts, e.g., when interested practitioners are specialized in control theory, but not in software engineering.

\subsection{Contributions}
\IEEEpubidadjcol
To address the monolithic nature of current controllers, we analyze the common responsibilities that are frequently duplicated across implementations:\begin{itemize}
  \item \textit{Reference acquisition and validation.} Controllers often accept references via ROS2 topics \cite{ros2-topic} and actions \cite{ros2-action} and verify that they are dimensionally consistent, numerically valid, and within expected limits.
  \item \textit{Real-time-safe handling.} Many controllers receive external references asynchronously but must compute a reference at the controller loop rate. 
  This requires handling the communication between non-real-time and real-time contexts.
  \item \textit{Interpolation.} Planners typically produce waypoints at rates differing from the low-level controller frequency, necessitating internal interpolation logic.
\end{itemize}
Decoupling those from control law calculation is possible and would make new controllers easier to implement and more reusable across different applications.

To reach this goal, we propose a \texttt{ros2\_control} software module called \emph{Reference Generator} (RG) that encapsulates all of the responsibilities cited above.
RGs handle reference acquisition via standard ROS2 communication primitives and produce single-point references at the controller's sampling period.
A key design choice is to exclusively use the controller chaining mechanism as the communication interface between RGs and downstream controllers \cite{ros2-controller-chaining}.
By leveraging ROS2 interfaces, this mechanism promotes cross-compatibility among modules developed by different organizations while keeping the design extensible for future needs.
Consequently, new controllers developed within this framework are agnostic to how references are acquired, making the implementation of the control law their sole responsibility.
This design also makes RGs backward-compatible with any ROS2 controller that supports chaining.

Specifically, this paper makes the following contributions:
\begin{enumerate}
  \item We design a novel architecture that separates reference processing (acquisition, validation, and interpolation) from the control law algorithm within the \texttt{ros2\_control} ecosystem.
  \item We provide two implementations for \emph{Reference Generator}: a \textit{Joint-Space Reference Generator} (JRG) and a \textit{Task-Space Reference Generator} (TRG) for handling joint-space and Cartesian references, respectively.
  \item Additionally, we deliver a set of new controllers, to further demonstrate how our reference generators favor the creation of control chains: (i) PD with gravity compensation, (ii) Cartesian pose \cite{Chiacchio1991}, and (iii) admittance controllers. The latter is tailored to work with Cartesian references instead of joint-space ones, differently from the original \texttt{AdmittanceController} proposed in \texttt{ros2\_controllers}~\cite{ros2-controllers}.
\end{enumerate}
The proposed \emph{Reference Generator} and the new controllers are successfully demonstrated in various pipeline configurations on simulated and real Universal Robots \cite{universal-robots} UR10 and Franka Emika Robot \cite{haddadin_franka_2022,franka-emika-robot} (FER) manipulators.

\subsection{Control laws}\label{subs:control_laws}
Consider the robot dynamics \cite{featherstone_dynamics_2016}:
\begin{equation}\label{eq:robots_dynamics}
    \bm B(\bm q)\,\ddot{\bm q} + \bm C(\bm q,\dot{\bm q})\,\dot{\bm q} + \bm f(\dot{\bm q}) + \bm g(\bm q) = \bm \tau,
\end{equation}
where $\bm B$ is the inertia matrix, $\bm C$ collects Coriolis and centrifugal terms, $\bm f$ models friction, $\bm g$ the gravitational effects, and $\bm \tau$ are the actuation torques. 
Vectors $\bm q,\dot{\bm q},\ddot{\bm q}$ denote joint positions, velocities, and accelerations. 
The end-effector pose, twist and accelerations are denoted with $\bm x,\bm v,\dot{\bm v}$, respectively.

Here we recall the control laws used in the pipelines of Section \ref{sec:example_demos}.
\begin{itemize}
  \item \textbf{PD control with gravity compensation (PDGC):}
  \begin{equation}
    \bm \tau = \bm K_P(\bm q_{d}-\bm q) - \bm K_D\,\dot{\bm q} + \bm g(\bm q),
    \label{eq:gcpd}
  \end{equation}
  where $\bm K_P,\bm K_D \succ 0$ are symmetric positive-definite gain matrices, and $\bm q_{d}$ is the desired joint position profile~\cite{Kelly2005}.

  \item \textbf{PID control:}
  \begin{equation}
    \bm \tau = \bm K_P(\bm q_{d}-\bm q) + \bm K_D(\dot{\bm q}_{d}-\dot{\bm q}) + \bm K_I \int_{t_0}^{t} (\bm q_{d}-\bm q)\,\mathrm{d}t,
    \label{eq:pid}
  \end{equation}
  with integral gain matrix $\bm K_I \succ 0$ and desired joint velocities $\dot{\bm q}_{d}$\cite{Kelly2005-PID}.

  \item \textbf{Admittance control (AC):}
  The desired relationship between the commanded and measured wrenches is defined in task space as a mass-spring-damper system acting on the task error $\tilde{\bm x} = \bm x_d - \bm x_c$, with $\bm x_d$ and $\bm x_c$ denoting the desired and commanded end-effector poses, respectively:
  \begin{equation}
    \bm M_d\,\ddot{\tilde{\bm x}} + \bm K_D\,\dot{\tilde{\bm x}} + \bm K_P\,\tilde{\bm x} = \bm h_d - \bm h_e,
    \label{eq:admittance_dynamics}
  \end{equation}
  where $\bm M_d, \bm K_D, \bm K_P \succ 0$ are diagonal positive-definite matrices specifying the desired dynamics, $\bm h_e$ is the measured external wrench, and $\bm h_d$ is the commanded feedforward wrench. 
  The vectors $\dot{\tilde{\bm x}}$ and $\ddot{\tilde{\bm x}}$ denote the first and second time derivatives of $\tilde{\bm x}$.  
  The resulting control law providing the commanded task motion is
  \begin{equation}
    \ddot{\tilde{\bm x}} = \bm M_d^{-1}\big(\bm h_d - \bm h_e - \bm K_D\,\dot{\tilde{\bm x}} - \bm K_P\,\tilde{\bm x}\big),
    \label{eq:admittance_control_law_final}
  \end{equation}
  whose integration yields the commanded pose $\bm x_c = \bm x_d - \tilde{\bm x}$ tracked by the downstream controller~\cite{Newman1992, Villani2016}.

  \item \textbf{Cartesian pose control (CPC):}
  \begin{equation}
    \dot{\bm q}_c = \bm J^{\dagger}(\bm q)\big(\bm K_P(\bm x_{d}-\bm x) + \bm v_{d}\big),
    \label{eq:cartesian_pose_control}
  \end{equation}
  where $\bm J^{\dagger}$ denotes a damped least-squares pseudoinverse of the geometric Jacobian $\bm J$, and $\dot{\bm q}_c$ is the resulting joint-space command~\cite{Chiacchio1991}. When only a position controller is available, $\dot{\bm q}_c$ is integrated to obtain target joint positions.
\end{itemize}

\begin{figure}
  \centering
  \includegraphics[width=\columnwidth]{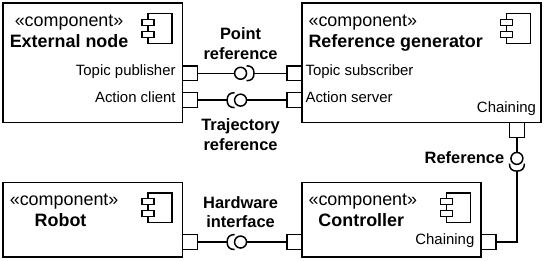}
  \caption{Component diagram. The \emph{Reference Generator} validates and interpolates incoming references, producing single, chainable references for the downstream controller.}
  \label{fig:component_diagram}
\end{figure}

\section{Design}\label{sec:design}
This section describes the proposed architecture and its behavior. 
An implementation is available in our repository\footnote{\url{https://github.com/unisa-acg/reference-generators-demo/tree/rap}}.

Fig.~\ref{fig:component_diagram} shows the main components and their interfaces.
In particular, we distinguish between two main components with complementary functionalities: the \textit{Reference Generator} and the \textit{Controller}.
The former:
\begin{itemize}
  \item receives single-point references or trajectories from external nodes;
  \item validates incoming references to ensure dimensional consistency, numerical correctness, and compliance with expected limits, and rejects invalid external references with appropriate error codes;
  \item performs interpolation of references to match the controller frequency;
  \item writes a single-point reference to the downstream controller via \texttt{ros2\_control}'s chaining mechanisms.
\end{itemize}
The latter, instead:
\begin{itemize}
  \item exposes a chainable controller interface according to \texttt{ros2\_control}'s chaining mechanisms;
  \item reads a preprocessed reference input from the chainable interfaces;
  \item executes the control law and performs hardware-interface or downstream-controller-reference writes;
\end{itemize}
In this novel design, controllers are responsible only for executing control laws and interacting with hardware/downstream controllers, while reference acquisition, validation, and interpolation are delegated to the RG. 
This separation avoids code duplication and enhances reusability within the \texttt{ros2\_control} framework.

\begin{figure}
    \centering
    \includegraphics[width=\linewidth]{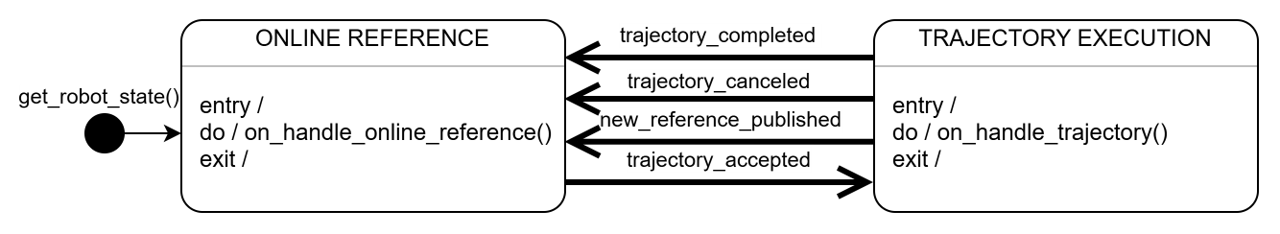}
    \caption{Finite state machine of the \emph{Reference Generator}.}
    \label{fig:rg_fsm}
\end{figure}

The RG operates in two complementary modes: forwarding online point references received on a topic, and executing trajectories received via actions. 
As shown in Fig.~\ref{fig:rg_fsm}, its behavior is governed by a finite state machine (FSM) with two states: \texttt{ONLINE\_REFERENCE} and \texttt{TRAJECTORY\_EXECUTION}.
On startup, if no reference has been published on the topic yet, the RG reads the current robot state and computes an initial reference.
In \texttt{ONLINE\_REFERENCE}, the generator forwards the most recent reference retrieved from the topic at the controller's sampling period, keeping it constant until a new message or trajectory arrives.
When a trajectory is received, the FSM switches to \texttt{TRAJECTORY\_EXECUTION} and the RG produces interpolated points at each control period according to trajectory timestamps. 
When the trajectory is completed, aborted, or a new reference is received from the topic, the FSM returns to \texttt{ONLINE\_REFERENCE}. 
If a new trajectory arrives while another is being executed, the RG aborts the current trajectory and starts the new one, providing action feedback and result codes accordingly. 
Importantly, all non-real-time processing (action/subscriber handling, parsing and validation) occurs outside the real-time controller path (thanks to the decoupling offered by the RGs), while writing to the chainable interface remains deterministic.

\section{Experimental validation}\label{sec:example_demos}
\subsection{Experimental setup and results}\label{subsec:experimental_setup_and_results}
The experiments validate the design presented in Section~\ref{sec:design}. 
In particular, across all experiments, the goal is to demonstrate three key properties of the proposed architecture:
\begin{enumerate}
    \item \textbf{Controller-agnosticity}: RGs can easily be configured to work with downstream controllers.
    \item \textbf{Robot-agnosticity}: the same control pipeline can be exercised on different hardware platforms and Gazebo \cite{koenig_design_2004} simulations.
    \item \textbf{Reference interpolation}: the JRG and TRG preprocess incoming references and reliably deliver single-point references at the controller's sampling period.
\end{enumerate}

  \begin{figure}
      \newcommand{\imagewidth}{0.32\columnwidth}
      \centering
      \includegraphics[width=\imagewidth]{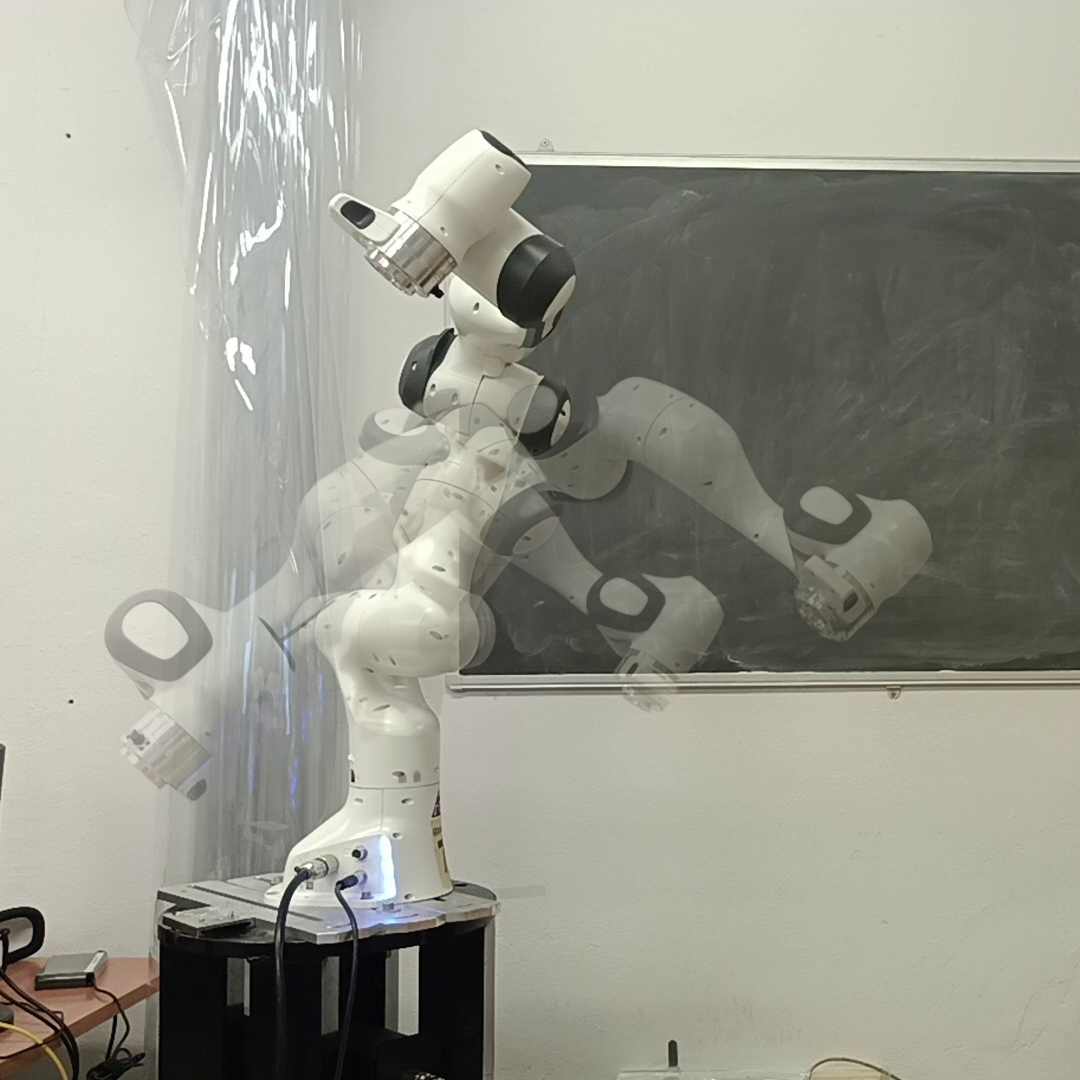}
      \hfill
      \includegraphics[width=\imagewidth]{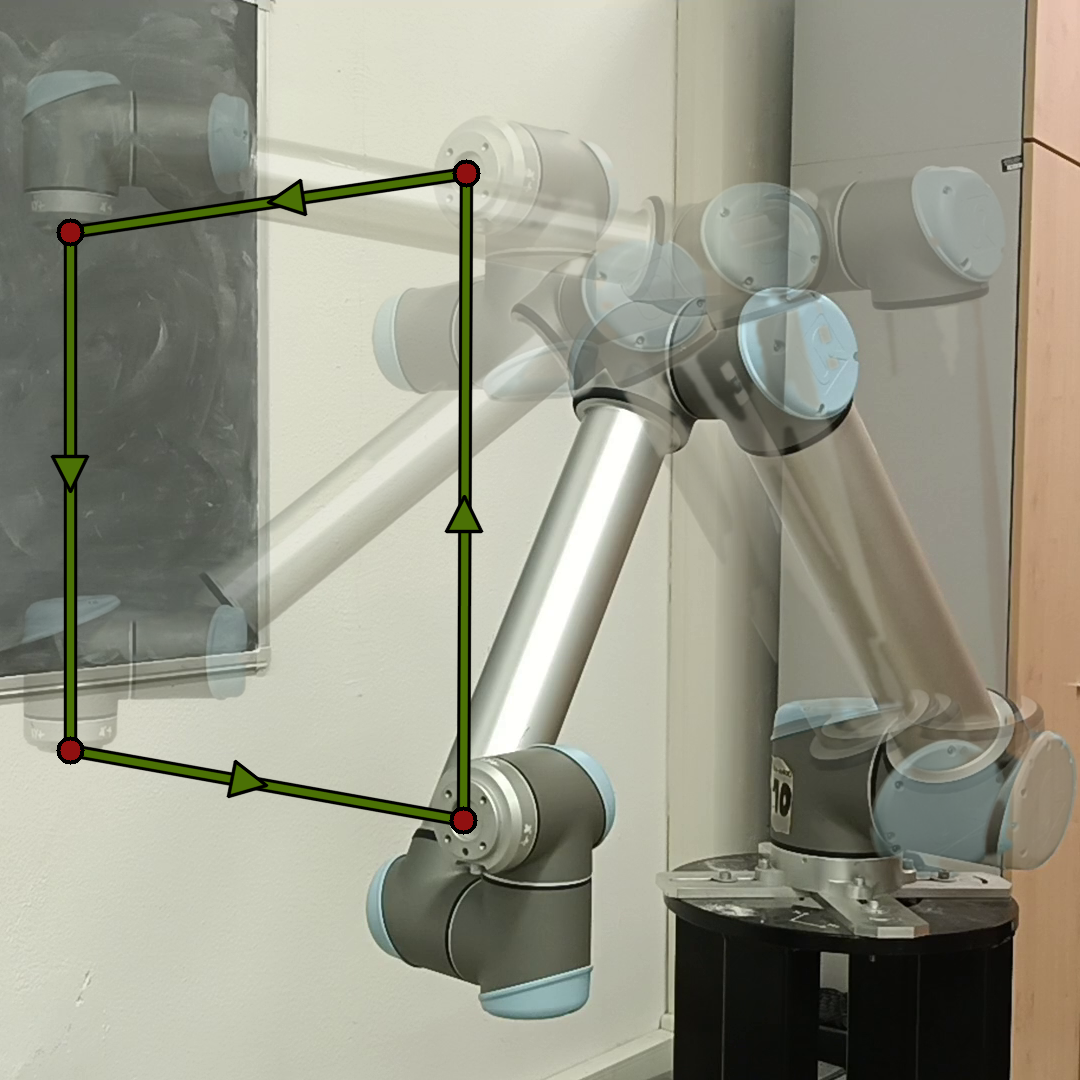}
      \hfill
      \includegraphics[width=\imagewidth]{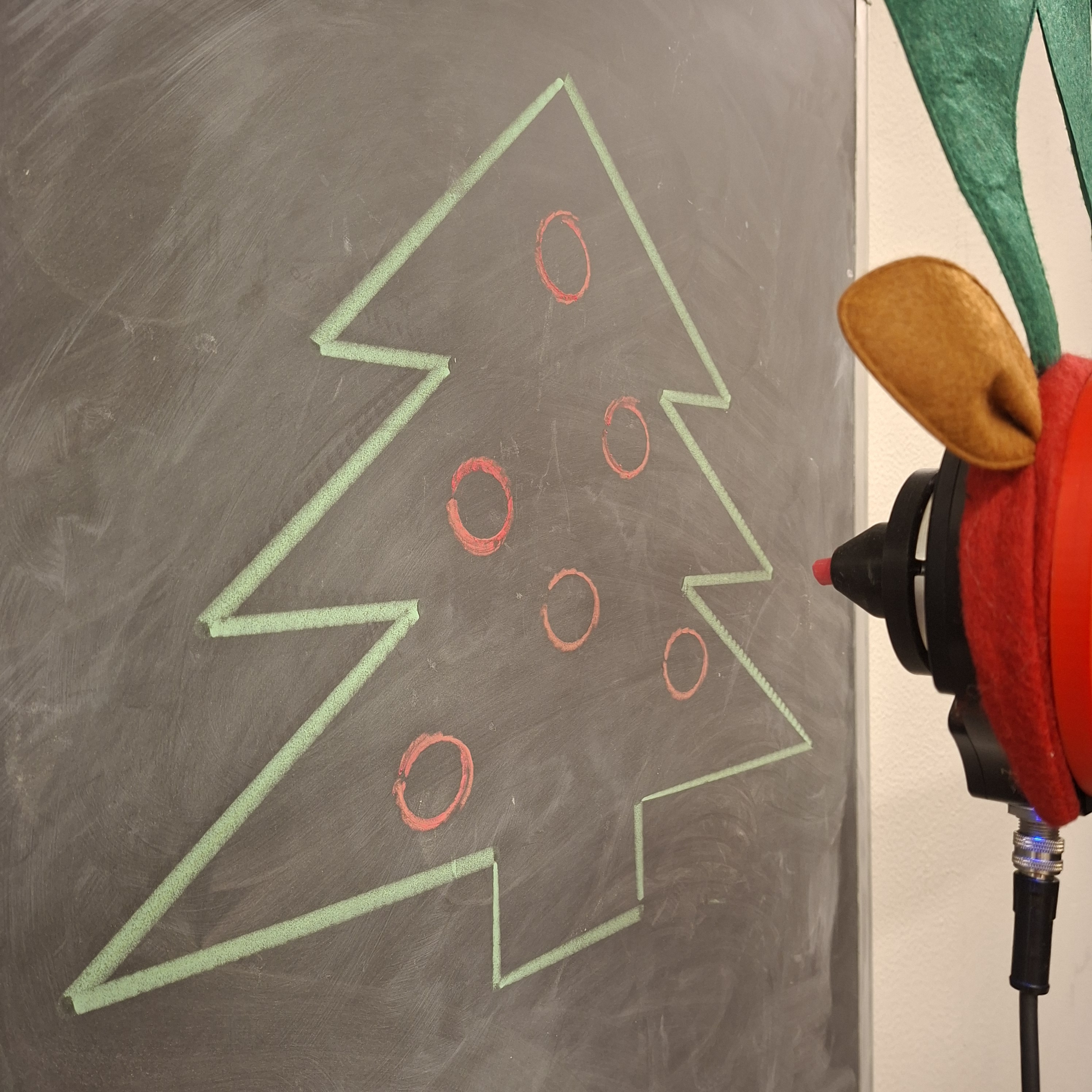}
      \caption{Experiments on real robots: (left) FER executing an excitation trajectory in the JRG \textrightarrow{} PDGC experiment; (center) control points defining the square Cartesian trajectory in the TRG \textrightarrow{} CPC experiment on the UR10; (right) UR10 drawing on a blackboard with the TRG \textrightarrow{} AC \textrightarrow{} CPC pipeline.}
      \label{fig:ur10_square_trajectory}
  \end{figure}
  \begin{figure}
      \centering
      \includegraphics[width=0.8\columnwidth]{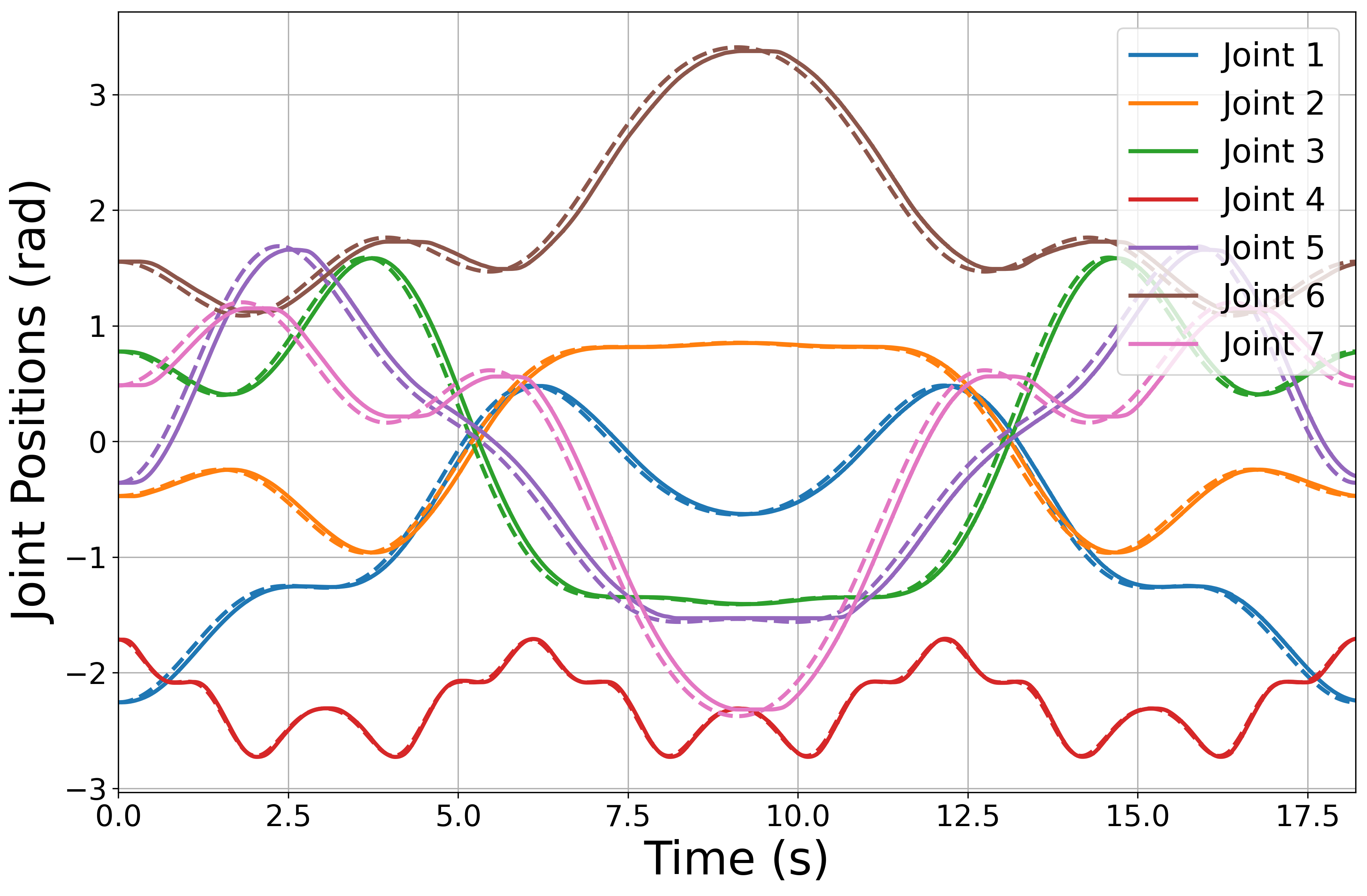}
      \caption{Reference (dashed) and measured (solid) joint positions on the real FER under PD control with gravity compensation.}
      \label{fig:gravity_compensation_pd_real_franka}
  \end{figure}

\begin{figure}
      \centering
      \includegraphics[width=0.8\columnwidth]{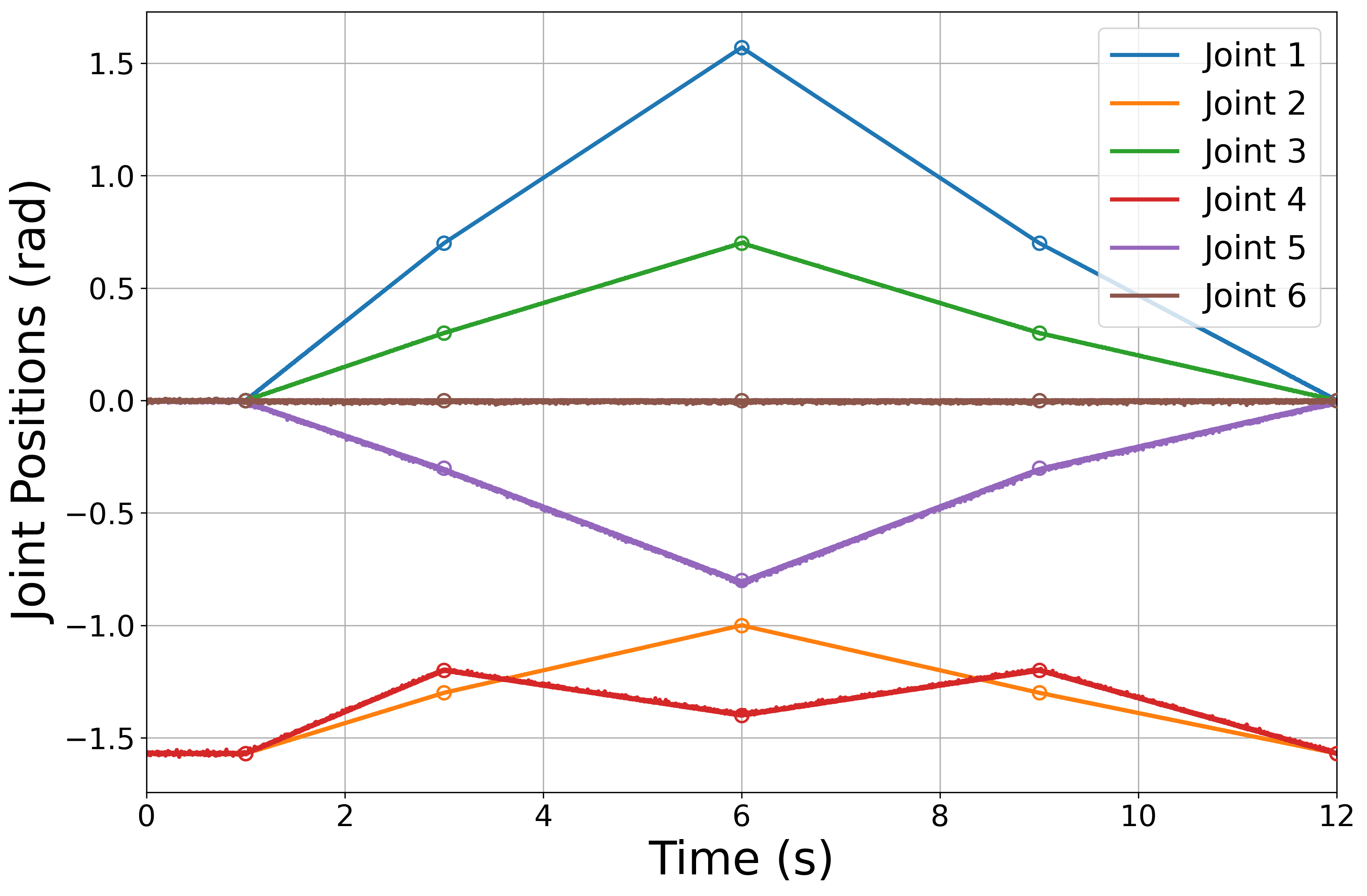}
      \caption{Reference (circles) and measured (lines) joint positions on the simulated UR10 under PID control.}
      \label{fig:pid_controller_simulated_ur10}
  \end{figure}

\begin{figure}
      \centering
      \includegraphics[width=0.50\columnwidth]{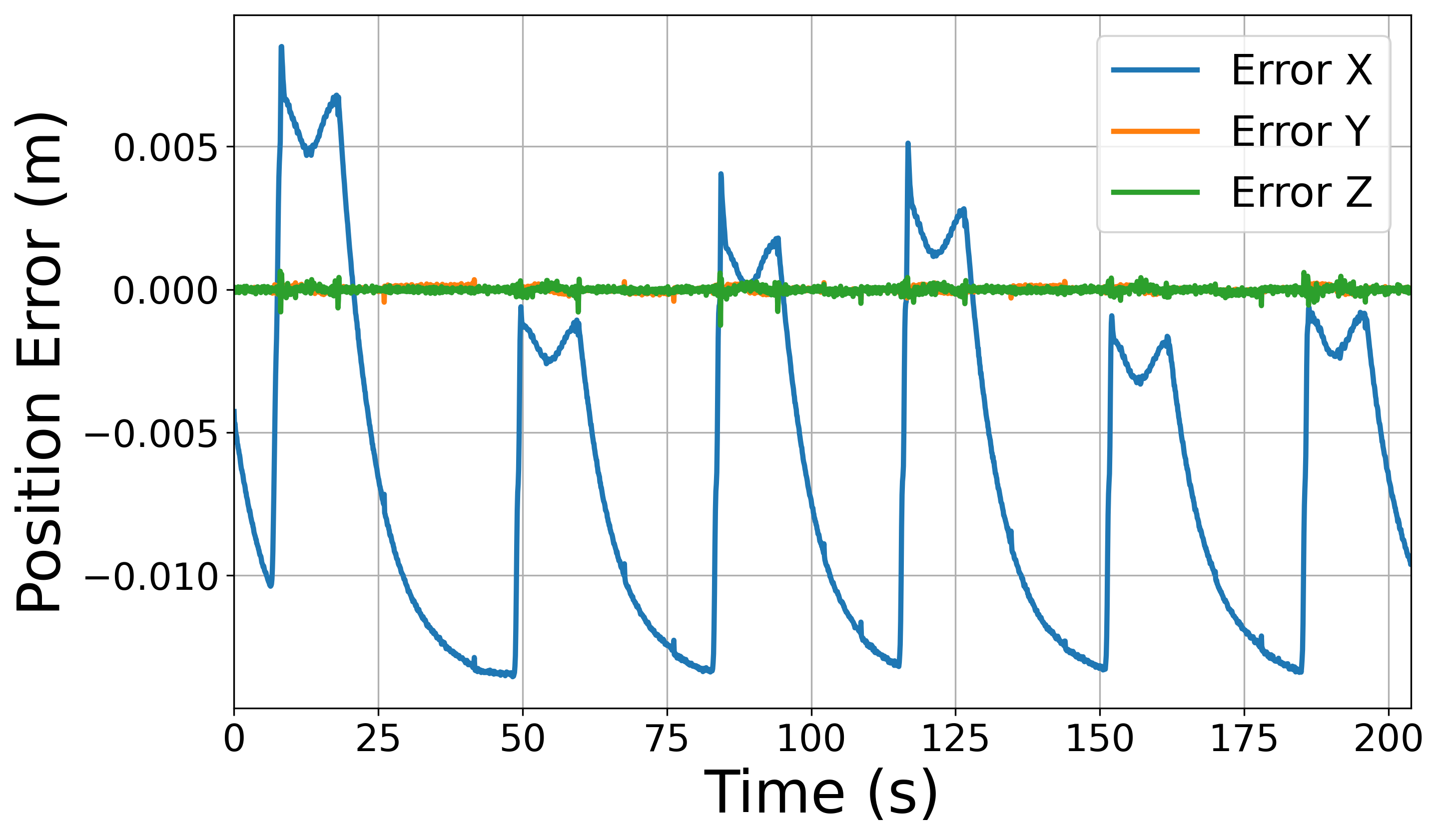}%
    \hfill
      \includegraphics[width=0.50\columnwidth]{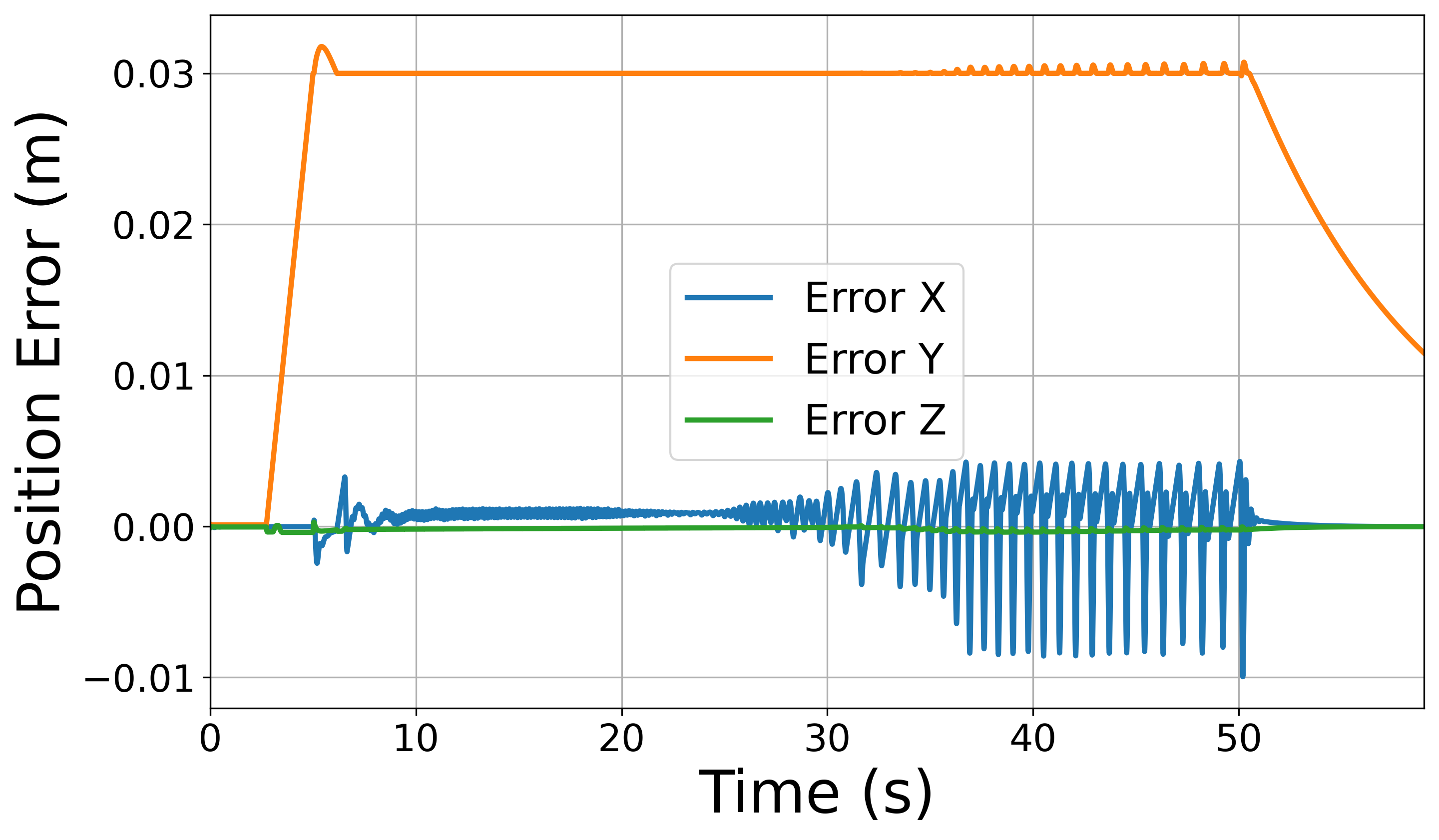}%
    \caption{End-effector position errors on the real (left) and simulated (right) UR10 under admittance control. X and Y axes are contact directions in base frame, respectively.}
    \label{fig:admittance_errors_combined}
  \end{figure}

Each experiment pairs a RG with one or more downstream controllers. 
Since the single-point reference functionality of the RG trivially consists of forwarding the received topic reference to the downstream controller, the experiments focus on trajectory tracking via actions.
The effectiveness of the design and of the controller pipelines is qualitatively demonstrated through trajectory-tracking results.
The following control pipelines are tested:
\begin{itemize}
  \item \textbf{JRG \textrightarrow{} PDGC \textrightarrow{} simulated FER.}  
  The JRG provides $\bm q_{d}$ to the PDGC \eqref{eq:gcpd} at frequency $f = \SI{1}{\kilo\hertz}$, which issues torque commands to the simulated FER.
  \item \textbf{JRG \textrightarrow{} PDGC \textrightarrow{} real FER.}  
  The same pipeline is deployed on the real FER. 
  Fig.~\ref{fig:ur10_square_trajectory} shows the robot executing the trajectory, with tracking results reported in Fig.~\ref{fig:gravity_compensation_pd_real_franka}.
  In this experiment, the user provides joint-space waypoints at the same rate as the controller ($f = \SI{1}{\kilo\hertz}$), so the interpolation functionality of the JRG is not exercised.
  \item \textbf{JRG \textrightarrow{} PID \textrightarrow{} simulated UR10.}  
  The user provides a few sparse joint-space waypoints to the JRG via action, which interpolates them at the controller rate ($f = \SI{1}{\kilo\hertz}$) and sends them to the PID controller \eqref{eq:pid} in simulation.
  Tracking results are shown in Fig.~\ref{fig:pid_controller_simulated_ur10}.
  \item \textbf{TRG \textrightarrow{} CPC \textrightarrow{} simulated UR10.}  
  The user sends sparse Cartesian waypoints to the TRG via action; the TRG provides Cartesian pose setpoints to the CPC at $f = \SI{1}{\kilo\hertz}$, which computes joint targets via inverse kinematics \eqref{eq:cartesian_pose_control} and sends them to the simulated UR10.
  \item \textbf{TRG \textrightarrow{} CPC \textrightarrow{} real UR10.}  
  Similar pipeline on the real UR10, but the frequency is reduced to $f = \SI{125}{\hertz}$ to accommodate hardware limitations.
  An illustration of the trajectory waypoints given by the user is shown in Fig.~\ref{fig:ur10_square_trajectory}.
  Tracking results (after interpolation) are in Fig.~\ref{fig:cartesian_pose_controller_real_ur10}.
  \item \textbf{TRG \textrightarrow{} AC \textrightarrow{} CPC \textrightarrow{} real UR10.}  
  The user sends a few Cartesian waypoints to perform an interaction experiment with the real UR10, consisting of sliding a piece of chalk on a blackboard, as shown in Fig.~\ref{fig:ur10_square_trajectory}. 
  The TRG provides Cartesian setpoints at $f = \SI{125}{\hertz}$, which the AC modulates according to the measured end-effector wrench, as explained in \eqref{eq:admittance_dynamics} and \eqref{eq:admittance_control_law_final}; the modulated setpoints are executed by the CPC on the real UR10. The actual robot behavior is shown in Fig.~\ref{fig:admittance_errors_combined} in terms of position errors.
  \item \textbf{TRG \textrightarrow{} AC \textrightarrow{} CPC \textrightarrow{} PID \textrightarrow{} simulated UR10.}  
  A composite pipeline consisting of TRG, AC, CPC, and PID controllers is executed on the simulated UR10 in an interaction scenario. The trajectory consists of a few Cartesian waypoints for a wall-sliding motion.
  In this experiment, the TRG provides Cartesian setpoints at $f = \SI{1}{\kilo\hertz}$, which the AC modulates according to the measured end-effector wrench; the modulated setpoints are converted to joint targets by the CPC and finally into effort commands by the PID controller before being sent to the simulated UR10. 
  Fig.~\ref{fig:admittance_errors_combined} shows the position error between the TRG references and the actual end-effector position over time.
\end{itemize}

\subsection{Discussion}

The experiments confirm the three key properties stated in Section~\ref{subsec:experimental_setup_and_results}.
The same implementations of the RGs were paired with different downstream controllers (PDGC, PID, CPC, AC) without modifying the code of any components. 
All controllers successfully received the preprocessed references and executed their control laws.
Identical control pipelines were executed on both simulation and real hardware (FER and UR10) without architectural changes, demonstrating that the design separates reference generation and controller logic from robot-specific details.

\begin{figure}
      \centering
      \includegraphics[width=0.48\columnwidth,height=0.28\textheight,keepaspectratio]{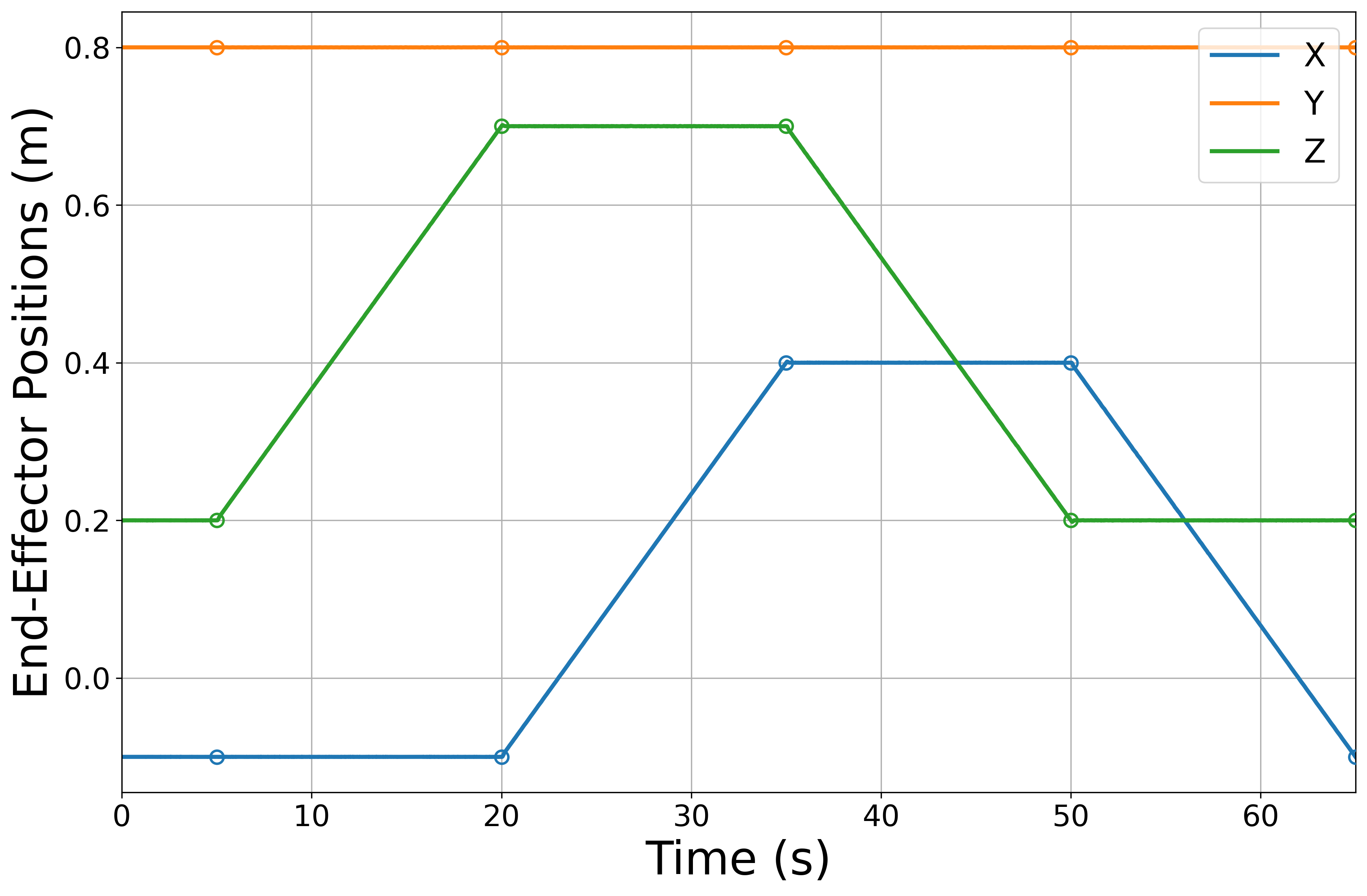}
      \hfill
      \includegraphics[width=0.48\columnwidth,height=0.28\textheight,keepaspectratio]{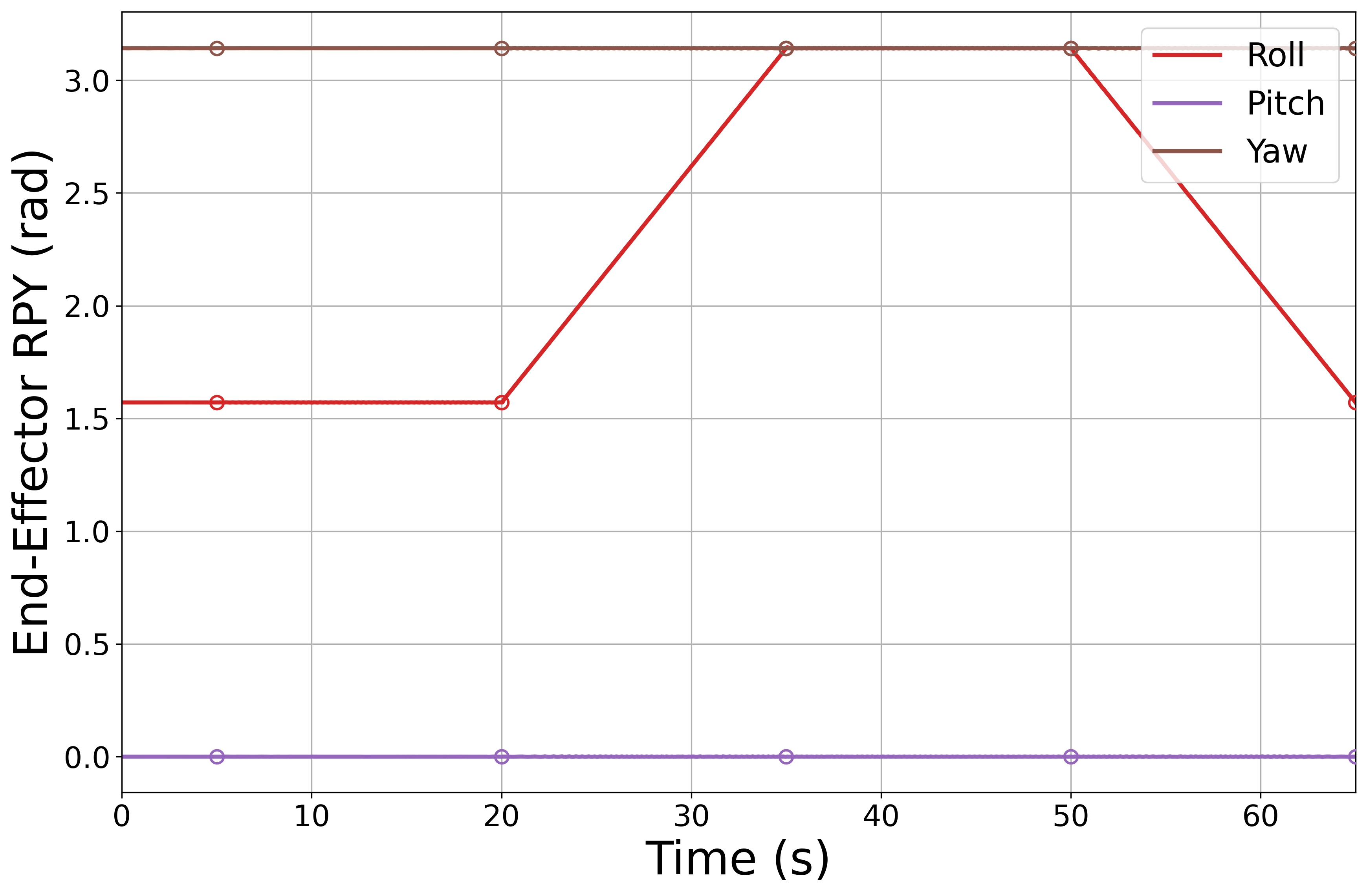}
      \caption{Reference (circles) and measured (lines) end-effector position (left) and orientation (right) signals on the real UR10 under Cartesian pose control.}
      
      \label{fig:cartesian_pose_controller_real_ur10}
  \end{figure}

Both TRG and JRG correctly preprocessed sparse, time-stamped waypoints and produced single-point references at the controller sampling period (see Fig.~\ref{fig:pid_controller_simulated_ur10} and~\ref{fig:cartesian_pose_controller_real_ur10}).
Larger errors observed in the admittance experiments are expected, since interaction forces intentionally modulate the motion during contact, yielding compliant behavior.

The results also confirm that multi-layer pipelines (e.g., TRG \textrightarrow{} AC \textrightarrow{} CPC \textrightarrow{} PID) can be assembled without specialized architectural design, demonstrating the framework's flexibility and support for hierarchical control. 
Finally, both ROS2-native (PID) and new proposed controllers (PDGC, CPC, and AC) were successfully validated, confirming their compatibility with the proposed design.

To quantify the reduction in code complexity enabled by the proposed architecture, we compare against the \texttt{JointTrajectoryController} \footnote{\url{https://github.com/ros-controls/ros2_controllers/tree/humble}}.
This controller can be regarded as a monolithic coupling of a JRG with a PID control law, which is precisely the design this work proposes to separate by delegating reference management to the RGs, leaving the controllers with only control-law code.
We counted approximately 905 of 1,745 source lines (52\%) dedicated to reference handling features: trajectory validation, action-server and subscriber callbacks, goal management, trajectory-point processing, and real-time safe buffering.
By following the SRP~\cite{Martin2008}, the RGs are a one-time investment shared across all downstream controllers: any new controller does not need to re-implement the aforementioned features, yielding compounding savings as the number of controllers grows.


\section{Conclusions}\label{sec:conclusions}

This paper presented a modular architecture for ROS2 that separates reference processing from control-law execution, realized through two RGs (JRG and TRG).
The work demonstrated that this separation improves reusability, simplifies controller design, and enables modular construction of flexible control pipelines through chaining.

The JRG and TRG were shown to be compatible not only with the custom controllers introduced in this work, but also with the \texttt{ros2\_control} PID controller, showing that the same RGs can be used with controllers developed by the community as long as they adhere to the \texttt{ros2\_control}'s chaining mechanisms.

The pipelines defined in Section~\ref{sec:example_demos} were configured by only changing the parameters in the configuration file (e.g., interface names, control period, and gains), demonstrating the limited effort required for reconfiguration. 
In addition, the successful development of new control laws based on the same RGs shows that they provide a stable abstraction for extending and prototyping additional control strategies.

The architecture naturally allows for extensions at the reference generation level; while linear interpolation was sufficient for our demonstrations, the modular design easily accommodates higher-order interpolation (e.g., quintic splines) strategies to implement in future works, to improve smoothness and reference fidelity without affecting downstream controllers.

Furthermore, extending the TRG to produce velocity commands would also benefit mobile-base controllers --- e.g., the \texttt{DiffDriveController} --- by decoupling its subscriber logic and introducing trajectory-handling capabilities.

\appendix[Description of Supplementary Materials]

\begin{table*}
\centering
\caption{Description of Supplementary Materials}
\label{tab:supplementary-materials}
\renewcommand{\arraystretch}{2.0}
\begin{tabular}{p{0.17\textwidth}|p{0.78\textwidth}}
\textbf{Material} & \textbf{Description} \\ \hline
Video\newline
(rap-rg-video.mp4) &
The video shows the three experiments executed on the real robots:
\begin{enumerate*}
\item PDGC on the real FER;
\item CPC on the real UR10 robot;
\item AC $\rightarrow$ CPC pipeline on the real UR10 robot.
\end{enumerate*}
The video can be viewed on YouTube at \url{https://youtu.be/qpMYzO2Cpx8}
\\

Source code\newline
(online) &
The source code is hosted in a GitHub repository, containing the ROS2 packages implementing the reference generators (JRG and TRG) described in the paper, together with documentation and launch files to use them.
The readers can directly access the GitHub repository at \url{https://github.com/unisa-acg/reference-generators-demo/tree/rap}.\\

UML Class Diagram\newline
(rap-rg-class-diagram.png) &
The Class Diagram better illustrates the main \textit{Reference Generators} classes (\texttt{ReferenceGenerator}, \texttt{JointSpace\allowbreak ReferenceGenerator}, and \texttt{TaskSpaceReferenceGenerator}), their methods, and their relationships.
\end{tabular}
\end{table*}

This work is supported by the supplementary materials detailed in Table~\ref{tab:supplementary-materials}.

\bibliographystyle{IEEEtran}
\bibliography{rap-2025-reference-generator}

\newpage

\end{document}